\documentclass[sigconf]{acmart}

\AtBeginDocument{%
  }

\setcopyright{acmlicensed}
\copyrightyear{2024}
\acmYear{2024}
\acmDOI{XXXXXXX.XXXXXXX}

\acmConference[XXX '25]{Make sure to enter the correct
  conference title from your rights confirmation emai}{March XX--XXX,
  2024}{XXX, XXX}
\acmISBN{978-1-4503-XXXX-X/18/06}

\usepackage[linesnumbered,ruled,vlined]{algorithm2e}
\usepackage{algpseudocode} 

\SetCommentSty{mycommfont}
\SetKwInput{KwInput}{Input}                
\SetKwInput{KwOutput}{Output}  

\usepackage{tabularx}
\usepackage{multirow}
\usepackage{graphicx}
\usepackage{subfigure}
\usepackage{url}
\usepackage{hyperref}

\renewcommand{\paragraph}[1]{\noindent \textbf{#1}}

\begin{document}

\title{Can Self Supervision Rejuvenate Similarity-Based Link Prediction?}

\author{Chenhan Zhang}
\email{chenhan.zhang@student.uts.edu.au}
\affiliation{%
  \institution{University of Technology Sydney}
  \city{Sydney}
  \country{Australia}
}

\author{Weiqi Wang}
\email{weiqi.wang-2@student.uts.edu.au}
\affiliation{%
  \institution{University of Technology Sydney}
  \city{Sydney}
  \country{Australia}
}

\author{Zhiyi Tian}
\email{zhiyi.tian@student.uts.edu.au}
\affiliation{%
  \institution{University of Technology Sydney}
  \city{Sydney}
  \country{Australia}
}

\author{James J.Q. Yu}
\email{jqyu@ieee.org}
\affiliation{%
  \institution{Southern University of Science and Technology}
  \city{Shenzhen}
  \country{China}
}

\author{Dali Kaafar}
\email{dali.kaafar@mq.edu.au}
\affiliation{%
  \institution{Macquarie University}
  \city{Sydney}
  \country{Australia}
}

\author{An Liu}
\email{anliu@suda.edu.cn}
\affiliation{%
  \institution{Soochow University}
  \city{Suzhou}
  \country{China}
}

\author{Shui Yu}
\email{shui.yu@uts.edu.au}
\affiliation{%
  \institution{University of Technology Sydney}
  \city{Sydney}
  \country{Australia}
}


\begin{abstract}
Although recent advancements in end-to-end learning-based link prediction (LP) methods have shown remarkable capabilities, the significance of traditional similarity-based LP methods persists in unsupervised scenarios where there are no known link labels. 
However, the selection of node features for similarity computation in similarity-based LP can be challenging. Less informative node features can result in suboptimal LP performance.
To address these challenges, we integrate self-supervised graph learning techniques into similarity-based LP and propose a novel method: \textbf{S}elf-\textbf{S}upervised \textbf{S}imilarity-based \textbf{LP} (\textbf{3SLP}).
3SLP is suitable for the unsupervised condition of similarity-based LP without the assistance of known link labels.
Specifically, 3SLP introduces a dual-view contrastive node representation learning (DCNRL) with crafted data augmentation and node representation learning.
DCNRL is dedicated to developing more informative node representations, replacing the node attributes as inputs in the similarity-based LP backbone.
Extensive experiments over benchmark datasets demonstrate the salient improvement of 3SLP, outperforming the baseline of traditional similarity-based LP by up to $21.2\%$ (AUC). 
\end{abstract}

\ccsdesc[500]{Computing methodologies~Feature selection}


\keywords{Link prediction, Graph neural networks, Self-supervised learning}


\maketitle

\section{Introduction} \label{intro}

Link prediction (LP) is one of the most intriguing and enduring challenges in the field of graph mining, which involves predicting the probability of a link between two unconnected nodes based on available information in the graph, such as node attributes \cite{yang2022few}. 
LP contributes to a myriad of real-world applications
such as friend recommendation in social networks \cite{wang2017nodes}, drug-drug interaction prediction \cite{wang2023scientific}, and knowledge graph completion \cite{shi2018open}.

LP methods can be broadly classified into two categories: similarity-based LP and learning-based LP.
While learning-based LP has shown strong capabilities, there remains a significant role for similarity-based LP methods. 
On the one hand, they heavily rely on known link information as labels to supervise the learning process \cite{zhang2018link, zhu2020beyond}. However, in some realistic applications, link information can be absent, where we only have an \textit{edgeless graph}.
For example, in the ``cold start'' phase of the drug-drug interaction, where link labels may not be readily available \cite{huang2022coadti}. 
Similarity-based LP assists in hypothesis generation by suggesting potential interactions derived from the attributes of drugs, where these label information can be further utilized in unsupervised learning on the data.
On the other hand, the ``black box'' nature of widely adopted end-to-end neural network-based LP methods makes it difficult to gain insights into what inclusive modules or specific data features are pivotal in determining LP.
Comparatively, similarity-based is a transparent and intuitive method that is more conducive to the analysis of results.


\paragraph{Research Gap.}
However, similarity-based LP methods are heuristic due to their predefined similarity measures \cite{kumar2020link}. Consequently, the quality of input node features can significantly impact the final LP performance.
This study reexamines similarity-based LP and explores further advancement within the existing framework. 
We believe that the similarity-based LP methods can be enhanced by improving the representation of the original node attributes. 
Following this vein, a further hypothesis is that node-level features enriched with topological information can enhance prediction accuracy, considering that link existence depends on both individual node features and their relative positions and neighborhood structures.
Nevertheless, as previously mentioned, similarity-based LP nowadays is often used in contexts when there are no or limited known links to label information. This condition impedes the utilization of topological information.

\paragraph{Motivation.}
Typically, the effectiveness of similarity-based LP methods significantly hinges on the feature engineering of the original node attributes. 
In this regard, a key hypothesis is that node-level features enriched with topological information can enhance prediction accuracy, considering that link existence depends on both individual node features and their relative positions and neighborhood structures in the graph.
Traditional similarity-based LP methods either use original node attributes directly, or use the node features processed by feature selection or dimensionality reduction techniques \cite{kumar2020link}.
These approaches are suboptimal, as these attributes or features only incorporate individual information.
One method is that we can leverage graph representation learning (GRL) techniques \cite{jiao2022graph} to learn complex topological information and encode it into node features.
However, as previously mentioned, similarity-based LP is often involved in a context without link-label information (graph structural information), which impedes this process.
Self-supervised learning (SSL) has emerged as an effective solution for extracting representations from unlabeled data. To address the unsupervised challenges, SSL techniques employ data augmentation and pretext tasks to furnish specific supervision signals during the learning process, thereby enabling the learned representations to be expressive and can be applied across various downstream tasks \cite{chen2020simple,he2020momentum}. 
Recently, there have witnessed endeavors to advance SSL to existing GRL techniques \cite{qiu2020gcc}. 
Some of them have demonstrated success in capturing latent topological information and developing insightful node representations \cite{wang2023self,rong2020self}.
Inspired by these methods, leveraging SSL appears promising as a means to develop node-level features for similarity-based LP.

\begin{figure*}
    \centering
    \includegraphics[width=\linewidth]{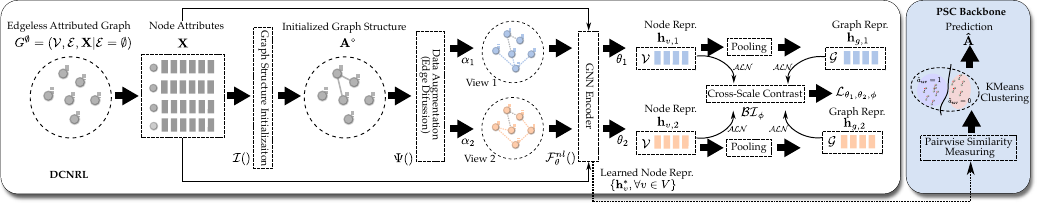}
    \caption{Schematic of 3SLP.}
    \label{fig:framework}
\end{figure*}

\paragraph{Contribution.}
In this paper, we propose a novel approach to improve similarity-based LP: \textbf{S}elf-\textbf{S}upervised \textbf{S}imilarity-Based \textbf{LP} (\textbf{3SLP}). 
Considering the capabilities of self-supervised learning (SSL) techniques in unsupervised graph representations learning \cite{rong2020self, wang2023self}, the main idea of 3SLP is to integrate SSL into node representation learning and use the developed node representations for similarity-based LP.
We first introduce a widely adopted pairwise similarity-based clustering (PSC) backbone \cite{he2021stealing,GraphMI} as the testbed. 
This backbone operates in an unsupervised manner and only requires node-level features as inputs to implement prediction, constituting a practical and commonly adopted scheme for similarity-based LP.
Then, we design a dual-view contrastive node representation learning (DCNRL) to develop informative node representations as inputs instead of original node attributes to the PSC backbone to prompt LP performance.
In DCNRL, we first address the absence of structural information by initializing a graph structure. 
Considering the significance of latent higher-order proximity in the node representation learning, we then employ graph diffusion to generate the two augmented views in the data augmentation phase. 
Furthermore, inspired by \cite{DGI,hassani2020contrastive}, we apply cross-scale contrasting on the extracted node-level and graph-level representations to develop more comprehensive node representations.
Our experimental results showcase the superiority of 3SLP, evidencing its ability to outperform the baseline by as much as $21.2\%$.
Our main contributions are summarized as follows:
\begin{itemize}
    \item We rejuvenate similarity-based LP problem and propose 3SLP, a novel approach that leverages SSL to develop informative node representation for improving similarity-based LP. We introduce a dual-view contrastive learning method (DCNRL) for the node representation development.
    \item We address the absence of links label information by graph structure initialization and edge diffusion. We design a cross-scale contrastive objective between node- and graph-level representations to develop comprehensive node features.
    \item We demonstrate a significant performance improvement in similarity-based LP contributed by our method. Code is available at \url{https://anonymous.4open.science/status/WSDM_25-EDB2}.
\end{itemize}

\paragraph{Roadmap.}
The remainder of the paper is organized as follows. Section \ref{sec:PD} presents preliminary knowledge about this study. In Section \ref{sec:method}, we elaborate on the proposed 3SLP method. We present the evaluation to demonstrate the effectiveness of the proposed scheme in Section \ref{sec:Experiment}. Section \ref{sec:conclusion} concludes the paper and discusses future work.

\section{Prelimiaries} \label{sec:PD}
In this section, we first introduce the investigated LP task. Then, we will provide some background information about graph representation learning. 
Lastly, we elucidate the differences between our study and contemporary LP methods.

\subsection{Usage Scenario of Similarity-Based LP} \label{NAOLP}
The link prediction (LP) task is to determine whether a given pair of nodes are linked (i.e., an edge exists) \cite{kumar2020link}.
Let $G = (\mathcal{V},\mathcal{E},\mathbf{X})$ be an attributed graph with node set $\mathcal{V} \in \mathbb{R}^{\vert \mathcal{V} \vert}$, edge set $\mathcal{E}$, and node attribute matrix $\mathbf{X} = \{\mathbf{x}_{i}\} \in \mathbb{R}^{\vert \mathcal{V} \vert \times d}$ where $d$ is the dimension of the node attributes. The adjacency matrix of $G$ is denoted by $\mathbf{A} = \{a_{ij}\} \in \mathbb{R}^{\vert \mathcal{V} \vert \times \vert \mathcal{V} \vert}$ where entry $a_{ij} = 1$ if edge $(i,j) \in \mathcal{E}$ and $a_{ij} = 0$ otherwise. 
Let $\mathcal{E}^{k}$ and $\mathcal{E}^{r}$ represent the known and real edge sets of a graph, respectively.
To replicate scenarios typical of similarity-based LP use cases, where only node attributes may be accessible without link information (such as in the initial stages of drug-drug interactions), our study particularly considers edgeless attributed graphs.
An edgeless attributed graph can be defined as $G^{\emptyset    } = (\mathcal{V},\mathcal{E}^{k},\mathbf{X} \vert \mathcal{E}^{k} = \emptyset)$. 
Therefore, the objective of similarity-based LP on an edgeless attributed graph can be concisely expressed as: $\mathbf{X} \rightarrow \mathcal{E}^{r}$.
Notably, the investigated case offers a harsh testing ground for similarity-based LP methods; they need to demonstrate the ability to efficiently discern and extract relevant information from potentially subtle differences in attributes to achieve accurate predictions.

\subsection{GNN-Based Graph Representation Learning}
Graph neural networks (GNNs) are extensively utilized to develop node- and graph-level representations. Take message-passing-based GNNs as an example, a canonical node representation (embedding) encoding process includes (1) $\mathcal{AGG}()$: aggregate messages received from neighbor nodes; (2) $\mathcal{UPD}()$: update the node representation by non-linear transformation \cite{niepert2016learning}. Let $\mathcal{N}(v)$ be the set of $1$-hop neighbor nodes of node $v$, the process flow can be expressed as:
\begin{equation}
\small
 \mathcal{F}^{nl}_{\theta}(\mathbf{X}, \mathbf{A}): \ \mathbf{h}_v^l =\mathcal{UPD}\left(\mathbf{h}_v^l, \mathcal{AGG}\left(\left\{\mathbf{h}_u^{l-1}, \forall u \in \mathcal{N}(v)\right\}\right)\right), \label{eq:node_encode}
\end{equation}
where $\mathbf{h}_v^{l} \in \mathbb{R}^{h}$ denotes the node-level representation of node $v$ at layer $l$ and $h$ is the dimension. 
On this basis, to develop graph-level representation $\mathbf{h}_g \in \mathbb{R}^{h}$, a graph pooling operation (e.g., max pooling and mean pooling) can be performed on all node embeddings:
\begin{equation}
\small
 \mathcal{F}^{gl}(\mathbf{h}_{v},\forall v \in V): \  \mathbf{h}_g = \mathcal{POOL}(\mathbf{h}_{v},\forall v \in V).  \label{eq:graph_encode}
\end{equation}

\subsection{Relationship between Our Study and Contemporary LP Studies}
While there is a broader range of advanced end-to-end learning-based LP methods \cite{mavromatis2020graph, ahn2021variational, ahn2021variational, pan2021neural, zhu2021neural, guo2023linkless}, which lead the mainstream and state-of-the-art in the LP field, our study does not focus on improving or surpassing these approaches. Instead, our main motivation is to enhance classic similarity-based LP methods, given their practicality in certain use cases. Moreover, as our study is orthogonal to end-to-end learning-based LP methods, our method can further assist these approaches, such as by generating reasonable pseudo labels during cold-start phases. 

Furthermore, it is worth mentioning that we do not focus on improving similarity-based LP methods from the perspective of metrics (e.g., similarity score computation) or algorithms (e.g., clustering algorithms). Our primary focus is on enhancing the input node features/attributes. In this context, the employed backbone (as introduced in Section \ref{sec:dmc}) for the similarity-based LP is constructed based on widely evaluated and adopted similarity-based LP methods \cite{kumar2020link, he2021stealing, chen2023survey}.

\section{Our Method} \label{sec:method}
\subsection{Method Overview}
In this section, we introduce our proposed 3SLP. The core architecture of 3SLP is two-fold: (1) pairwise similarity-based clustering backbone (PSC) (empirical backbone) and (2) dual-view contrastive node representation learning (DCNRL) (our design) . 
Specifically, 3SLP first leverages DCNRL to develop informative node representations from the edgeless attributed graph. 
Then, the PSC backbone takes these node representations as input to predict the target links. A schematic view of 3SLP is given in Figure \ref{fig:framework}. The algorithmic details of 3SLP are given in Algorithm~\ref{alg:main}

\subsection{Pairwise Similarity-Based Clustering Backbone} \label{sec:dmc}
We first introduce pairwise similarity-based clustering (PSC) based on the previous construction in the literature \cite{he2021stealing,GraphMI}.
PSC forms a concise yet effective unsupervised clustering method, serving as the backbone of our method that implements similarity-based LP. 
The PSC backbone mainly incorporates two steps. 
The first step is node similarity measuring between pairwise input node features. For any two nodes $u$ and $v$, and their respective features $\mathbf{i}_{u}$ and $\mathbf{i}_{v}$, the node similarity measuring can be formulated as:
\begin{equation}
\small
    s_{uv} = \Upsilon(\mathbf{i}_{u}, \mathbf{i}_{v}), \label{eq:sim}
\end{equation}
where $\Upsilon$ represents the adopted similarity measurer. In this work, we involve five commonly-used symmetric similarity metrics: (1) cosine similarity; (2) cosine distance; (3) Euclidean distance; (4) Manhattan distance; and (5) correlation distance. By node similarity measuring, we can obtain $|\mathcal{V}| \times |\mathcal{V}|$ pairs of node representation similarity score, denoted as $\mathcal{S}$.
Next, PSC applies an unsupervised clustering algorithm to perform binary classification on the computed similarity scores. Specifically, we adopt the K-Means algorithm. The number of clusters is set to $2$, corresponding to positive (link) and negative (no link) node pairs. Treating it as a ranking problem here, PSC calculates the average similarity score for each cluster\,---\,based on the adopted metrics, the cluster whose node pairs with a lower (higher) average score is considered linked (not linked)\footnote{An exception is when cosine similarity metric is adopted where we consider the cluster with higher value the linked one.}. The clustering process can be written as:
\begin{equation}
\small
\begin{aligned}
& \ \ \mathcal{S}^*_1, \mathcal{S}^*_2 \leftarrow \operatorname{KMeans}(\{s_{uv} \in \mathcal{S}\}) \\
&\left\{\begin{array}{l}
\hat{a}_{uv} \in \mathcal{S}^*_1 = 1, \hat{a}_{uv} \in \mathcal{S}^*_2 = 0 \ \ \ \text{if} \ \mathcal{AVG}(\mathcal{S}^*_1) > \mathcal{AVG}(\mathcal{S}^*_2) \\
\hat{a}_{uv} \in \mathcal{S}^*_1 = 0, \hat{a}_{uv} \in \mathcal{S}^*_2 = 1 \ \ \ \text{else}. \end{array}\right.
\end{aligned} \label{eq:kmeans}
\end{equation}

Given that the PSC is primarily used as the testbed for our new designs, further refinement, such as balancing positive and negative samples (class imbalance issue), falls outside the scope of this study.
Building upon the PSC backbone, we can utilize the informative node feature to prompt LP. A naive approach directly takes the original node attribute information $\mathbf{x}$ as input (PSC-NA). 
In this work, we treat PSC-NA as the major baseline.



\begin{algorithm}[htbp]
\DontPrintSemicolon
\KwInput{edgeless attributed graph $G^{\emptyset    } = (\mathcal{V},\mathcal{E},\mathbf{X} \vert \mathcal{E} = \emptyset)$, initializer $\mathcal{I}$,  edge diffuser $\Psi$, teleportation probabilities $\alpha_1$ and $\alpha_2$, GNN encoder $\mathcal{F}^{nl}_{\theta}$, \# learning epochs $T$, learning rate $\eta$, similarity measurer $\Upsilon$}
\KwOutput{predicted links $\hat{\mathbf{A}}$}
\BlankLine
$\mathbf{A}^{\circ} \leftarrow \mathcal{I}(\mathbf{X})$ \tcp*[l]{initialize a graph structure}
$\Tilde{\mathbf{A}}^{\circ}_1 \leftarrow \Psi_{\alpha_1}(\mathbf{A}^{\circ})$ \tcp*[l]{augment view 1}
$\mathbf{A}^{\circ}_2  \leftarrow \Psi_{\alpha_2}(\mathbf{A}^{\circ})$ \tcp*[l]{augment view 2}
$\mathcal{F}^{nl}_{\theta1}, \mathcal{F}^{nl}_{\theta2} \leftarrow$ initialize two GNN encoders\;
\For{each round $t = 1,2, \ldots$, $t \in T$}{
$[\mathbf{h}_{v,1}, v \in \mathcal{V}] \leftarrow \mathcal{F}^{nl}_{\theta1}(\mathbf{X}, \mathbf{A}^{\circ}_1)$\;
$[\mathbf{h}_{v,2}, v \in \mathcal{V}] \leftarrow \mathcal{F}^{nl}_{\theta2}(\mathbf{X}, \mathbf{A}^{\circ}_2)$\;
\tcp*[l]{get node-level repr.}
$\mathbf{h}_{g,1} \leftarrow  \mathcal{F}^{gl}(\mathbf{h}_{v, 1},\forall v \in V)$\;
$\mathbf{h}_{g,2} \leftarrow  \mathcal{F}^{gl}(\mathbf{h}_{v, 2},\forall v \in V)$\;
\tcp*[l]{get graph-level repr.}
$\mathbf{H}_{v,1}, \mathbf{H}_{v,2}, \mathbf{H}_{g,1}, \mathbf{H}_{g,2} \leftarrow \mathcal{ALN}([\mathbf{h}_{v, 1}], [\mathbf{h}_{v, 2}]), \mathbf{h}_{g,1}, \mathbf{h}_{g,2})$\;
\tcp*[l]{dimension alignment}
$\mathcal{L} \leftarrow \min\limits_{\theta_1, \theta_2, \phi} \mathcal{BI}_{\phi} (\mathbf{H}_{g,1}, \mathbf{H}_{v,2} ;  \mathbf{\Check{H}}_{g,1}, \mathbf{\Check{H}}_{v,2} 
)+\mathcal{BI}_{\phi}(\mathbf{H}_{g,2}, \mathbf{H}_{v,1}  ; \mathbf{\Check{H}}_{g,2}, \mathbf{\Check{H}}_{v,1})$ via Eq. \eqref{eq:local_global}\;
$(\theta1, \theta2, \phi) \leftarrow (\theta1, \theta2, \phi) -\eta \cdot \nabla\mathcal{L}(\theta1, \theta2, \phi)$\;
\tcp*[l]{optimize cross-scale contrast objective}
}
$[\mathbf{h}^{*}_{v}] \leftarrow \frac{1}{2} \mathcal{F}^{nl}_{\theta1}(\Tilde{G}^{\circ}_1) + \mathcal{F}^{nl}_{\theta2}(\Tilde{G}^{\circ}_2))$\;
\tcp*[l]{get ss-learned node repr.} 
\For{each node pair $(u, v) \in G^{\emptyset}$}{$s_{uv} \in \mathcal{S} \leftarrow \Upsilon(\mathbf{h}^{*}_{u}, \mathbf{h}^{*}_{v})$}
\tcp*[l]{measure dist. between node repr.} 
$\mathcal{S}^*_1, \mathcal{S}^*_2 \leftarrow \operatorname{KMeans}([s_{uv} \in \mathcal{S}])$\;
\If{$\operatorname{AVG}(\mathcal{S}^*_1) > \operatorname{AVG}(\mathcal{S}^*_2)$}{
$1 \leftarrow \hat{a}_{uv} \in \mathcal{S}^*_1$, $0 \leftarrow \hat{a}_{uv} \in \mathcal{S}^*_2$\; 
}
\Else{$0 \leftarrow \hat{a}_{uv} \in \mathcal{S}^*_1$, $1 \leftarrow \hat{a}_{uv} \in \mathcal{S}^*_2$}
\tcp*[l]{cluster pari-wise dist.} 
$\hat{\mathbf{A}} \leftarrow ([\hat{a}_{uv}=1], [\hat{a}_{uv}=0])$\;
\caption{3SLP}
\label{alg:main}
\end{algorithm}

\subsection{Dual-View Contrast Node Representation Learning (DCNRL)}
Here, we introduce the proposed DCNRL method for generating informative node features for input into the PSC backbone. 
DCNRL employs a widely adopted dual-view contrastive learning framework. On this basis, we further introduce specifically tailored data augmentation and contrastive learning modules, catering to the unique needs of edgeless attributed graphs and similarity-based LP. 
The detailed designs of these components will be elaborated on in the following sections.

Within the edgeless attributed graph, no graph structural (i.e., links) information is known.  
We cannot use any known graph structural information for data augmentation and further node representation learning. Therefore, we propose first to initialize a graph structure.
Particularly, we adopt the cosine similarity metric to obtain the similarity between a node and all other nodes and select the nodes with top-$k$ nearest nodes to connect with.
This method resembles the creation of weak labels based on the attribute homophily \cite{yang2021diverse}.
Denote a graph initializer as $\mathcal{I}$, this step can be formulated as:
\begin{equation}
\small
    \mathbf{A}^{\circ} = \mathcal{I}(\mathbf{X})
\end{equation}
where $\mathbf{A}^{\circ}$ represents the adjacency matrix for the initialized graph structure.

However, the initialized graph structure may deviate significantly from the real graph structures, resulting in a noisy and insufficient configuration that may prove ineffectual for further GNN encoding.
It has been shown that \textit{diffusion} processes can enhance GNN performance on such noisy graph structures \cite{klicpera2019diffusion}.
Therefore, we employ edge diffusion as a data augmentation strategy on the initialized graph structure. This approach can imitate higher-order proximities among nodes, thus empowering GNN encoders to capture more intricate graph properties within the SSL framework. 
Specifically, we employ the Personalized PageRank (PPR) method for edge diffusion. \cite{berberidis2019node,li2019optimizing} have demonstrated PPR can smooth out the neighborhood over a noisy graph and restore more pertinent connections.
Given the adjacency matrix of the initialized graph $\mathbf{A}^{\circ}$ and diffuser $\Psi$, we can develop edge diffused graph structures using a closed-form expression for PPR derived from \cite{klicpera2019diffusion}, that is:
\begin{equation}
\small
\Tilde{\mathbf{A}}^{\circ}=\Psi(\mathbf{A}^{\circ}):=\alpha\left(\mathbf{I}-(1-\alpha) \mathbf{D}^{-1 / 2} \mathbf{A}^{\circ} \mathbf{D}^{-1 / 2}\right)^{-1},
\end{equation}
where $\mathbf{D}$ is the diagnonal degree matrix of $\mathbf{A}^{\circ}$ and $\alpha$ is the teleportation probability. Moreover, we apply two different teleportation probabilities $\alpha_1$ and $\alpha_2$ to the diffuser to create two distinct views:
\begin{equation}
\small
\begin{aligned}
\mathrm{VIEW 1}: \ \ \Tilde{\mathbf{A}}^{\circ}_{1}=\Psi_{\alpha_1}(\mathbf{A}^{\circ} ) \\
\mathrm{VIEW 2}: \ \ \Tilde{\mathbf{A}}^{\circ}_{2}=\Psi_{\alpha_2}(\mathbf{A}^{\circ}) 
\end{aligned}
\end{equation}
These two views, characterized by varying diffusion levels, furnish a progressive topological transition from a high-order perspective.
Consequently, the subsequent learning process is able to encode more comprehensive topological information within the node-level representation. 

For the LP purpose, we aim to develop node-level representations infused with topological information; however, these representations may fall short of being highly indicative due to the inherent limitations of the local aggregation function within a GNN encoder.
Drawing inspiration from \cite{hassani2020contrastive}, we propose to apply a cross-scale strategy during the contrasting phase. By juxtaposing features at different scales, we can extract pattern and structural information across multiple dimensions \cite{wu2021self}.  
This allows our GNN encoder to capture how the representation varies from different scales and thus learn a more robust representation.
Specifically, we first employ two respective GNN encoders for two views. While sharing the same architecture, the two encoders do not share parameters (with $\theta_1$ and $\theta_2$, respectively). 
Then, we use the GNN encoder to develop two scales  for each view progressively, namely, node-level representation and graph-level representation, respectively:
\begin{equation}
\small
\begin{aligned}
& \mathrm{VIEW 1}:\left\{\begin{array}{l}
\mathbf{H}_{v,1} = \mathcal{ALN}\left(\mathcal{F}_{\theta_1}^{nl}\left( \mathbf{X}, \Tilde{\mathbf{A}}^{\circ}_{1}\right)\right) \\
\mathbf{H}_{g,1} = \mathcal{ALN}\left(\mathcal{F}^{gl}\left( \mathbf{H}_{v,1} \right)\right)
\end{array}\right. \\
& \mathrm{VIEW 2}:\left\{\begin{array}{l}
\mathbf{H}_{v,2} = \mathcal{ALN}\left(\mathcal{F}_{\theta _2}^{nl}(\mathbf{X}, \Tilde{\mathbf{A}}^{\circ}_{2})\right)  \\
\mathbf{H}_{g,2} = \mathcal{ALN}\left(\mathcal{F}^{gl}\left( \mathbf{H}_{v,2} \right)\right),
\end{array}\right.
\end{aligned}
\end{equation}
where $\mathcal{ALN}$ represents the dimension alignment process to make each developed 
representation in the same dimension $\mathbf{H} \in \mathbb{R}^{\vert \mathcal{V} \vert \times h}$. The corresponding negative samples for each view are generated by our corruption function by randomly shuffling the node attributes $\mathcal{C}: \mathbf{X} \rightarrow \Tilde{\mathbf{X}}$ and we use $\mathbf{\Check{H}}_{v} = \mathcal{ALN}\left(\mathcal{F}_{\theta}^{nl}\left( \Tilde{\mathbf{X}}, \Tilde{\mathbf{A}}^{\circ}_{1}\right)\right) $ to denote the representation developed by $\Tilde{\mathbf{X}}$. Note that we do not choose to generate negative samples by corrupting the structural input $\Tilde{\mathbf{A}}^{\circ}$, as their included augmented edges are not the actual instances and thus may not render effective contrasting.

To better maximize the concordance between two positive views while minimizing that between positive and negative views, we compute the mutual information between their corresponding node-level and graph-level representations. Inspired by \cite{DGI}, 
we adopt neural network-based mutual information estimation, making this step more tractable. 
In this way, the final contrastive learning objective can be expressed as:
\begin{equation}
\resizebox{0.96\linewidth}{!}{$
    \min\limits_{\theta_1, \theta_2, \phi} \mathcal{BI}_{\phi} (\underbrace{\mathbf{H}_{g,1}, \mathbf{H}_{v,2}}_{\textbf{positive}} ;  \underbrace{\mathbf{\Check{H}}_{g,1}, \mathbf{\Check{H}}_{v,2}}_{\textbf{negative}} 
 )+\mathcal{BI}_{\phi}(\underbrace{\mathbf{H}_{g,2}, \mathbf{H}_{v,1}}_{\textbf{positive}}  ; \underbrace{\mathbf{\Check{H}}_{g,2}, \mathbf{\Check{H}}_{v,1}}_{\textbf{negative}}), \label{eq:local_global}
$}
\end{equation}
where $\mathcal{BI}_{\phi}$ represents a bilinear scoring function with shared parameters between the two views.
The promise of the design is that we can obtain \textit{globally relevant} node-level representations, capturing information across the entire graph. Such node representations are designed to preserve similarity for all node pairs, even for those that are distant \cite{donnat2018learning,DGI}.
Subsequently, to obtain the final node representation that will be used as input to the PSC backbone, we compute the average of the learned node representations derived from the two encoders, which can be expressed as:
\begin{equation}  
\small
\{\mathbf{h}^*_{v},\forall v \in V\}=\frac{1}{2}(\mathcal{F}_{\theta^{*}_1}^{nl}\left( \mathbf{X}, \Tilde{\mathbf{A}}^{\circ}_{1}\right) + \mathcal{F}_{\theta^{*}_2}^{nl}\left( \mathbf{X}, \Tilde{\mathbf{A}}^{\circ}_{2}\right)).
\end{equation}

\subsection{Complexity Analysis} \label{app:complex}
\paragraph{Time Complexity}
For time complexity, the main concern comes from the data augmentation and clustering steps that involve calculating pairwise node attribute and representation similarity, respectively. The naive approach to this calculation would have a time complexity of $O(N^2)$, where $N$ is the number of data points. However, it is possible to perform this step efficiently using parallelizable and approximate methods. For instance, NN-Descent \cite{dong2011efficient} utilizes MapReduce, which enables them to achieve approximate K-nearest neighbor (KNN) graphs in $O(n^{1.14})$ time complexity.

\paragraph{Space Complexity}
For space complexity, our method incorporates parameters from the two GNN encoders $f_{\theta_1}$ and $f_{\theta_2}$ and each one typically has $O(LNd + Ld^{2})$ (GCN) space complexity where $L$ is the layer number. The shareable MLP projector $\phi$ has a space complexity $O(d')$ linear to the representation dimension $d', d' \leq d$. The overall space complexity is $O(2LNd + 2Ld^{2} + d')$.

\begin{table}[h]
\small
    \centering
    \setlength{\tabcolsep}{3pt}
    \begin{tabular}{l c c c c c }
    \toprule
     \centering
         Dataset & Type &  \# Node & \# Edge & \# Node Attribute   \\ 
         \midrule
        Cora  & Citation & $2708$ & $5278$ & $1433$ \\
        Citeseer & Citation & $3312$ & $4536$ & $3703$ \\
        PubMed & Citation & $19717$ & $44338$ & $500$ \\
        Reddit & Posts & $1207$ & $179826$ & $602$ \\
        \bottomrule
    \end{tabular}
    \caption{Dataset statistics.}
    \label{tab:data_statistic}
\end{table}

\begin{table*}[t]
\small
\centering
\begin{tabular}{l@{\extracolsep{\fill}}ccccccccccc}
\toprule
\multirow{2.4}{*}{Method} & \multirow{2.4}{*}{Aux. Kn.} & \multirow{2.4}{*}{Metric} & \multicolumn{3}{c}{Cora} & \multicolumn{3}{c}{Citeseer} & \multicolumn{3}{c}{PubMed} \\
\cmidrule(r){4-6} \cmidrule(r){7-9} \cmidrule(r){10-12} &  & &  AUC (\%)  & AP (\%) &  $\Delta$ & AUC (\%)   &  AP (\%)  & $\Delta$ & AUC (\%)   &  AP (\%)  & $\Delta$  \\
    \midrule
    \multirow{3.5}{*}{PSC-NA} & \multirow{3.5}{*}{-}   & Cos sim. & $61.3\pm0.1$  & $61.3\pm0.1$ & - & $64.6\pm0.1$   & $64.7\pm0.1$ & - & $\mathbf{79.7}\pm\mathbf{0.1}$ & $\mathbf{75.6}\pm\mathbf{0.1}$ & - \\
    \multirow{3.9}{*}{(Major baseline)} & & Cos dis. & $61.3\pm0.1$  & $61.3\pm0.1$ & - & $64.6\pm0.1$    & $64.6\pm0.1$ & - & $79.7\pm0.1$  & $75.4\pm0.1$ & - \\
   &  & Eucl. & $57.3\pm3.6$  & $56.5\pm4.4$ & -  & $63.7\pm0.1$    & $63.6\pm0.1$ & - & $61.7\pm7.5$  & $61.2\pm3.2$ & - \\
   & & Corr. & $61.3\pm0.1$  & $61.3\pm0.1$ & - & $64.5\pm0.1$    & $64.5\pm0.1$ & - & $79.5\pm0.2$  & $76.0\pm0.1$ & - \\
    \midrule
    \multirow{3.5}{*}{PSC-Repr}  & \multirow{3.5}{*}{\checkmark} & Cos sim. & $63.3\pm1.1$  & $66.5\pm2.4$ & - & $64.7\pm3.8$   & $58.7\pm2.7$ & - & $50.2 \pm 0.1$ & $50.0\pm 0.1$ & - \\
    \multirow{3.9}{*}{(Zhang \textit{et al.} \cite{zhang2022inference})} & & Cos dis. & $66.3\pm5.2$  & $67.9\pm1.4$ & - & $60.0\pm1.0$    & $55.5\pm0.7$ & - & $50.2\pm0.1$  & $50.0\pm0.1$ & - \\
    & & Eucl. & $64.2\pm1.2$  & $60.2\pm0.6$ & -  & $65.7\pm0.9$    & $59.4\pm0.9$ & - & $34.5\pm0.7$  & $43.8\pm0.1$ & - \\
    & & Corr. & $74.3\pm2.8$  & $67.8\pm2.3$ & - & $65.8\pm2.5$    & $59.4\pm1.8$ & - & $50.2\pm0.1$  & $50.1\pm0.1$ & - \\
    \midrule
    \multirow{3.5}{*}{PSC-Pos}  & \multirow{3.5}{*}{\checkmark} & Cos sim. & $74.6\pm2.6$  & $66.5\pm2.4$ & - & $70.8\pm1.5$   & $63.1\pm1.2$ & - & $50.2\pm 0.1$ & $50.1\pm 0.1$ & - \\
    \multirow{3.9}{*}{(He \textit{et al.} \cite{he2021stealing})} & & Cos dis. & $76.2\pm1.5$  & $67.9\pm1.4$ & - & $71.6\pm1.3$    & $63.8\pm1.0$ & - & $50.2\pm0.1$  & $50.1\pm0.1$ & - \\
    & & Eucl. & $66.6\pm0.4$  & $60.2\pm0.5$ & -  & $66.7\pm0.2$    & $60.1\pm0.1$ & - & $34.0\pm0.5$  & $43.8\pm1.6$ & - \\
    & & Corr. & $76.0\pm2.5$  & $67.8\pm2.3$ & - & $72.1\pm2.8$    & $64.1\pm2.4$ & - & $50.2\pm0.1$  & $50.1\pm0.1$ & - \\
    \midrule
    \multirow{3.5}{*}{\textbf{3SLP}}  & \multirow{3.5}{*}{-} & Cos sim. & $\mathbf{77.1}\pm\mathbf{1.2}$  & $\mathbf{70.9}\pm\mathbf{1.7}$ & $17.0$ $\uparrow$& $\mathbf{84.0}\pm\mathbf{1.6}$    & $\mathbf{76.4}\pm\mathbf{1.7}$ & $19.4$ $\uparrow$ & $79.4\pm1.5$  & $73.3\pm1.1$ & $0.3$ $\downarrow$ \\
    \multirow{3.9}{*}{\textbf{(ours)}} & & Cos dis. & $\mathbf{78.7}\pm\mathbf{0.2}$  & $\mathbf{71.3}\pm\mathbf{0.1}$ & $17.4$ $\uparrow$ & $\mathbf{85.8}\pm\mathbf{0.9}$    & $\mathbf{81.7}\pm\mathbf{1.2}$ & $21.2$ $\uparrow$ & $\mathbf{80.5}\pm\mathbf{2.0}$  & $\mathbf{76.3}\pm2.5$ & $0.8$ $\uparrow$ \\
    & & Eucl. & $\mathbf{76.7}\pm\mathbf{0.3}$  & $\mathbf{70.0}\pm\mathbf{0.2}$ & $19.4$ $\uparrow$ & $\mathbf{82.3}\pm\mathbf{1.1}$    & $\mathbf{75.8}\pm\mathbf{1.9}$ & $18.6$ $\uparrow$ & $\mathbf{76.2}\pm\mathbf{0.1}$  & $\mathbf{70.4}\pm\mathbf{0.5}$ & $14.5$ $\uparrow$  \\
    & & Corr. & $\mathbf{78.6}\pm\mathbf{0.6}$  & $\mathbf{71.0}\pm\mathbf{0.4}$ & $17.3$ $\uparrow$ & $\mathbf{84.6}\pm\mathbf{0.8}$    & $\mathbf{77.3}\pm\mathbf{1.1}$ & $20.1$ $\uparrow$ & $\mathbf{79.6\pm2.3}$  & $\mathbf{76.3\pm2.5}$ & $0.1$ $\uparrow$\\
\bottomrule
\end{tabular}
    \caption{Performance comparison. The best results are highlighted in \textbf{bold}. ``Aux. Kn.'' is short for ``auxiliary knowledge''. $\Delta$ represents the performance improvements (AUC) from PSC-NA by 3SLP.}
\label{tab:NAO-LP}
\end{table*}

\begin{table}[ht]
\small
\centering
\setlength{\tabcolsep}{1mm}
\begin{tabular}{llccc}
\toprule
 & \multirow{2}{*}{{{{LP AUC (\%)}}}}  &  \multicolumn{3}{c}{{Target}} \\ 
 & & Cora & Citeseer & PubMed   \\  \cmidrule(r){3-5}
 \multirow{3}{*}{\rotatebox{90}{{Source}}} & Cora &
            $89.7\pm0.9$ & $51.0\pm1.5$ & $58.1\pm6.0$ \\
 & Citeseer & $53.8\pm6.5$ & $88.0\pm2.6$ & $62.9\pm4.4$  \\
 & PubMed & $57.0\pm4.2$ & $50.5\pm6.3$ & $93.2\pm2.4$ 
\\ \cmidrule(r){2-5}
 & Trans. Avg. & $55.4\pm5.4$ & $50.8\pm3.9$ & $60.5\pm5.2$ \\
 & \textbf{3SLP} & $\mathbf{73.9\pm2.6}$ & $\mathbf{85.8\pm0.9}$ & $\mathbf{80.5\pm2.0}$  \\
\bottomrule
\end{tabular}
\caption{LP Performance comparison with the dataset knowledge transfer-based GAE. The diagonal (in shadow) shows the non-transfer results of GAE. }
\label{table:domains}
\end{table}


\section{Evaluation} \label{sec:Experiment}
This section is structured as follows: First, we describe the experimental setup, detailing the datasets and models used. Next, we present the results of our experiments, followed by an analysis of graph homophily within the datasets. Finally, we conduct a hyperparameter test to evaluate the impact of different configurations on model performance.
\subsection{Experimental Setup}
\paragraph{Datasets:} We include four real-world and widely-adopted graph datasets, namely, Cora \cite{sen2008collective}, Citeseer \cite{sen2008collective}, PubMed \cite{namata2012query}, and Reddit \cite{hamilton2017inductive} in our experiments. The statistical information of the four datasets is summarized in Table \ref{tab:data_statistic}. 

\paragraph{Model and Parameter Settings.} 
We adopt single-layer GCN \cite{KipfGCN} as the default GNN encoder, and the hidden size is set to $512$. We use Adam \cite{adam} as the optimizer for the SSL where we have the number of training epoch $T=200$ and the learning rate $\eta=0.001$. 
In the graph initialization, we set $k=5$ for the top-$k$ nearest algorithm. 
The teleportation probabilities for the two diffusers are set to $\alpha_1=0.2$ and $\alpha_1=0.4$. We consider cosine distance as the default similarity metric of PNRSC.
The optimal values for some of these hyperparameters are discussed later. Experiments for each setting are run repeatedly five times to eliminate randomness.

\paragraph{Metrics.} We adopt area under the \textit{ROC curve (AUC)}, and \textit{average precision (AP)} as our metrics, which is widely adopted to measure the performance of binary classification in a range of thresholds \cite{ying2018hierarchical,chen2021machine,zhang2022inference}. We primarily refer to AUC as did in \cite{zhang2018link,he2021stealing}.

Additionally, we introduce two metrics for the graph homophily analysis. They are:

\textit{Attribute Assortativity Coefficient (AAC):}
\begin{equation}
    r_\text{aac}=\frac{\operatorname{Tr} (\mathbf{e}) -\left\|\mathbf{e}^2\right\|}{1-\left\|\mathbf{e}^2\right\|} \label{eq:aac}
\end{equation}
where $\mathbf{e}$ denotes the joint probability distribution of the specified attribute.

\textit{Degree Assortativity Coefficient (DAC):}
\begin{equation}
r_\text{dac}=\frac{\sum_{x y} x y\left(e_{x y}-a_x b_y\right)}{\sigma_a \sigma_b} \label{eq:dac}
\end{equation}
where $e_{x y}$ is the fraction of all edges in the graph that join together nodes with degree values $x$ and $y$, $a_x$ and $b_y$ are the fractions of edges that start and end at nodes with degree values $x$ and $y$ \cite{newman2003mixing}, and $\sigma_a $ and $\sigma_b$ are the standard deviations of the distributions $a_x$ and $b_y$.

\begin{figure}
    \centering
    \includegraphics[width=0.9\linewidth]{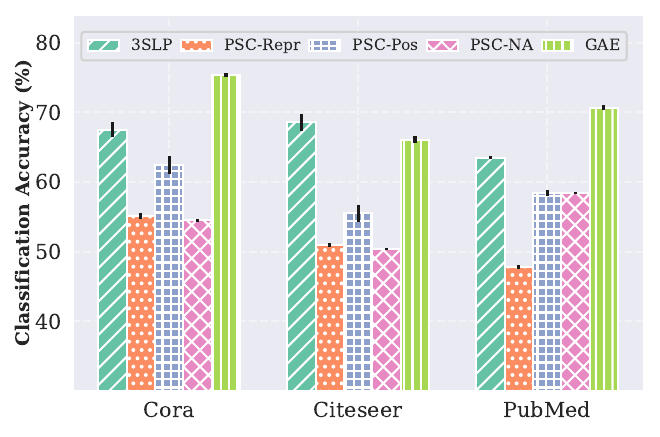} 
    \caption{Utility of predicted links on node classification tasks.}
    \label{fig:ex_knowledge}
\end{figure}


\begin{table}[ht]
\small
\centering
\resizebox{.95\linewidth}{!}{
\setlength{\tabcolsep}{2mm}
\begin{tabular}{lcccc}
\toprule
 GNN Encoder  & Cora & $\Delta$ & Citeseer & $\Delta$  \\ \cmidrule(r){1-1} \cmidrule(r){2-3} \cmidrule(r){4-5} 
 GCN (default) & $73.9\pm2.6$ & $12.6$ $\uparrow$ & $\mathbf{85.8\pm0.9}$ & $21.2$ $\uparrow$  \\
 GIN & $71.3\pm1.4$ & $10.0$ $\uparrow$ & $82.6\pm1.2$ & $18.0$ $\uparrow$  \\
 SGC & $\mathbf{74.1\pm0.5}$ & $12.8$ $\uparrow$ & $74.4\pm5.7$ & $9.8$ $\uparrow$  \\
 GraphSAGE & $67.0\pm4.9$ & $5.7$ $\uparrow$ & $63.4\pm8.0$ & $1.2$ $\downarrow$  \\
\bottomrule
\end{tabular}}
\caption{Comparison of LP performance (AUC (\%)) among 3SLP configuring different GNN encoders. $\Delta$ represents the improvements measured from the baseline.}
\label{table:gnn}
\end{table}


\begin{figure}
    \centering
    \includegraphics[width=1\linewidth]{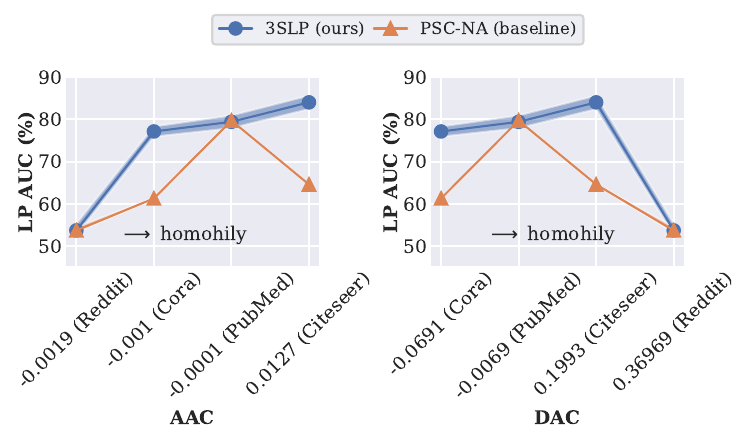}
    \caption{Graph assortativity coefficients versus LP performance (using cosine similarity in PSC). AAC measures graph's attribute homophily. DAC measures graph's topology homophily.}
    \label{fig:ex_aac}
\end{figure}

\begin{figure}
    \centering
    \includegraphics[width=\linewidth]{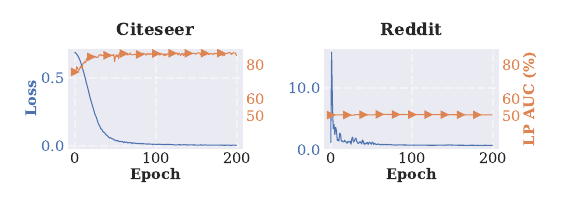} 
    \caption{Learning performance comparison between datasets Citesser (homophilic) and Reddit (heterophilic). Training loss per epoch (blue) and validation performance per epoch (orange).} 
    \label{fig:curves}
\end{figure}


\begin{figure*}[t]
    \centering
    \includegraphics[]{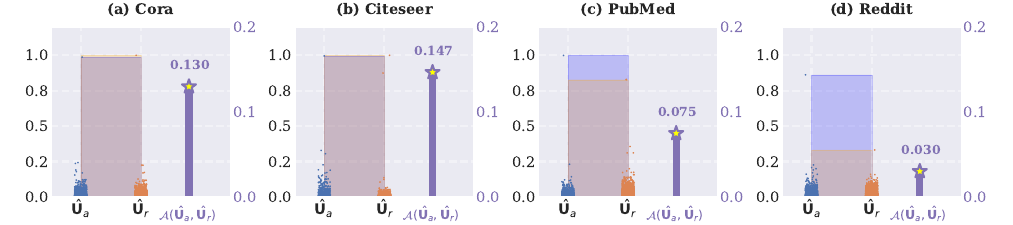}
    \caption{Spectrum alignment of adjacency matrix $\mathbf{A}$ and relation matrix $\mathbf{R}$. The realms of $\hat{\mathbf{U}}_{a}$ and $\hat{\mathbf{U}}_{r}$ are illustrated with blue and orange shadows, respectively. $\mathcal{A}(\hat{\mathbf{U}}_a, \hat{\mathbf{U}}_r)$ numerically indicates the level of spectrum alignment.}
    \label{fig:spectrum}
\end{figure*}

\begin{figure*}[t]
    \centering
    \includegraphics[width=0.95\textwidth]{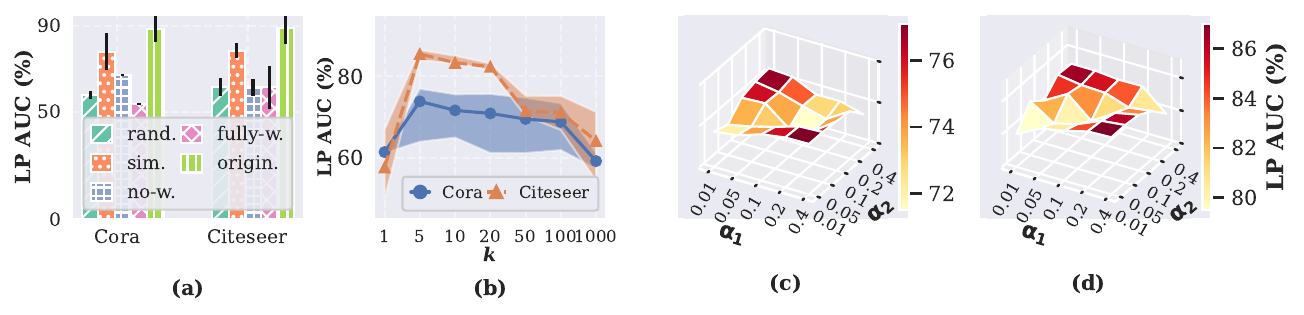} 
    \vspace{-1em}
    \caption{Ablation and hyperparameter tests. (a): Influence of graph structure initialization. (b): Influence of $k$ in similarity wiring. (c) and (d): Influence different teleportation probability $\alpha$ on Cora and Citeseer respectively.} 
    \label{fig:ex_ablation}
\end{figure*}

\subsection{Results}
\paragraph{Comparison with Baselines.}
We first compare the LP performance of the proposed method 3SLP with the baselines. 
As aforementioned, we consider pairwise node attribute similarity-based clustering (PSC-NA) the major baseline. 
From the results shown in Table \ref{tab:NAO-LP}, we can observe that 3SLP pronouncedly surpasses the baseline in most cases. For example, within the same PSC metric, 3SLP can exceed PSC-NA by $10.2\%$ (AUC) on the Cora dataset and $21.2\%$ on the Citeseer dataset.
Through self-supervision, 3SLP can develop node features that are more contributive than original node attributes with respect to the LP within the PSC backbone. However, we also notice that the performance improvement by our method on the PubMed dataset is not as pronounced as the other two datasets. We recognize that this pertains to graph homophily, which will be discussed later.

Additionally, we compare our method with the method involving auxiliary knowledge within the PSC backbone: (1) node representations (PSC-Repr) \cite{zhang2022inference}: the node representations developed from the trained GNN on node classification tasks in a supervised manner; (2) node posteriors (PSC-Pos) \cite{he2021stealing}: the node prediction posteriors in node classification tasks developed from the trained GNN in a supervised manner.
As shown in Table \ref{tab:NAO-LP}, 3SLP markedly outperforms these methods, implying the contribution of its developed node representations.

\paragraph{Comparison with Knowledge Transfer-Based Methods.}
Knowledge transfer is an effective technique for handling unsupervised tasks. We hereby compare 3SLP with the method of knowledge transfer between datasets, and specifically, we adopt a parameter sharing-based knowledge transfer method \cite{sun2020adashare}.
We adopt a Graph AutoEncoder (GAE) as the carrier of parameter sharing for knowledge transfer.
Each adopted dataset is considered a source and a target dataset, where we obtain a total of $9$ combinations.
Note that when the source and target datasets are the same, this situation becomes equivalent to regular supervised learning without knowledge transfer (see the diagonal in Table \ref{table:domains}).

The results in Table \ref{table:domains} indicate that 3SLP significantly outperforms other methods across all three datasets.
Specifically, it surpasses transferred GAE by an average of $19.1\%$ and $28.6\%$ on the Cora and Citeseer datasets, respectively. 3SLP even matches or exceeds the performance of non-transferred GAE in some scenarios. Notably, the effectiveness of transferred GAE is influenced by the source datasets used for training. The performance in the target domain is subject to variations depending on the source domain's quality and relevance, occasionally leading to negative transfer. 
In contrast, 3SLP leverages self-supervision to avoid the performance degradation commonly associated with knowledge transfer methods, thereby maintaining the integrity of the learning process.

\paragraph{Evaluating Predicted Links for Enhanced Node Classification in Cold Start Scenario}
Moreover, we wonder whether the predicted links possess good utility that can be employed for further analysis on the graph, particularly in cold start scenarios where limited information is available. To elucidate this, we examine whether these predicted links can be utilized to train an accurate GNN classifier for node classification tasks. To this end, we train a GCN classifier using predicted links in a supervised manner with node labels and evaluate its performance on a test set. Meanwhile, we compare the results with those developed by other tasks transferred knowledge-based methods as aforementioned.
From the result shown in Figure \ref{fig:ex_knowledge}, we find the proposed method achieves satisfactory outcomes, demonstrating its efficacy in cold start situations. 
The classification accuracy results, when developed from 3SLP's predicted links, consistently surpass those achieved by other methods.
Notably, on Citeseer, the classification accuracy developed from our predicted graph structure even betters the one on the original graph structure. This result implies that 3SLP can construct informative links that can be effectively utilized for various purposes, including addressing the challenges of cold start assistance.

\subsection{Analysis on Graph Homophily}
\paragraph{Influence of Graph Homophily.} 
When handling LP tasks, it is imperative to circumvent the homophilous assumption. Thereby, we investigate the influence of graph homophily on 3SLP's performance. We introduce the AAC and the DAC, which measure the attribute homophily and topology homophily of the graph, respectively \cite{newman2003mixing}. Their defintions are given in Eq.~\eqref{eq:aac} and \eqref{eq:dac}.
We include the relatively heterophilic dataset, Reddit, in this test to facilitate a more pronounced comparison. Figure \ref{fig:ex_aac} depicts a noticeable positive correlation between the LP performance of 3SLP and AAC: the higher the AAC, the better the LP performance of 3SLP; however, it does not show a similar trend between the LP performance of 3SLP and DAC.
We further compare 3SLP's learning performance between the relatively homophilous dataset Citeseer and the heterophilic dataset Reddit (see Figure \ref{fig:curves}). We find that 3SLP fails to learn informative information for the LP task in the heterophilic dataset. These results imply a highly possible correlation between 3SLP's effectiveness and attribute homophily assumption, which we will further explore in future work.

\paragraph{More Insights of Graph Homophily from Spectrum Analysis.} 
\cite{balestriero2022contrastive} provides a spectral theory-based study on the efficacy of SSL learned embedding in addressing downstream tasks. In spectrum space, they highlighted the significance of spectrum alignment between the relation matrix $\mathbf{R}$ (i.e., the transformation matrix for augmenting views $\mathbf{X}'$: $\mathbf{X}'=\mathbf{R}\mathbf{X}$) and the target matrix $\mathbf{Y}$. 
In our study, LP's target is the real adjacency matrix $\mathbf{A}$. By singular value decomposition, we first obtain
left singular vectors of $\mathbf{A}$ and $\mathbf{R}$, namely, $\hat{\mathbf{U}}_a$ and $\hat{\mathbf{U}}_r$.

We then evaluate the spectrum alignment between $\hat{\mathbf{U}}_a$ and $\hat{\mathbf{U}}_r$  (denote the left singular vector $\mathbf{U}_a$ with values greater than $0$ as $\hat{\mathbf{U}}_a$, similarly hereinafter). The spectrum alignment is quantified using the cosine similarity between the two singular vectors, denoted as $\mathcal{A}(\hat{\mathbf{U}}_a, \hat{\mathbf{U}}_r)$.
We particularly compare the results of the homophilic dataset Citeseer with those of the heterophilic dataset Reddit.
The results are present in Figure \ref{fig:spectrum}.
We observe that the result of the Citeseer dataset demonstrates relatively more substantial spectrum alignment; this aligns with a better LP performance observed on it. 
In contrast, the Reddit dataset presents a much smaller alignment value, correlating with its inferior LP performance. 
Moreover, we find that on Citeseer, the column of $\hat{\mathbf{U}}_{r}$ can \textit{span} $\hat{\mathbf{U}}_a$; this pattern is consistent with the deductions in \cite{balestriero2022contrastive} about superior efficacy of SSL learned embedding in handling downstream tasks. However, these patterns do not happen to Reddit. 
The relation matrix depends considerably on our graph initialization method, which is fundamentally anchored in the homophily assumption. 
We posit that higher adherence to the homophily assumption in our target graphs may potentially facilitate improved performance.

\subsection{Hyperparameter and Ablation Study} \label{sec:ex_abla}
\paragraph{Influence of Structural Initialization Methods.}
We first examine the performance of 3SLP with four different initialization methods for data augmentation. The results are shown in Figure \ref{fig:ex_ablation}(a). Similarity wiring achieves better results compared to others. This is because similarity wiring initializes more semantically meaningful graph structures regarding node attributes, which benefits subsequent data augmentation and the learning process. 
Furthermore, we investigate the hyperparameter $k$ in the similarity wiring method.
For hyperparameter $k$ in similarity wiring, we further set $k \in \{1, 5, 10, 20, 50, 100, 1000\}$ to examine LP performance. Note that when $k=0$ and $k=|\mathcal{V}|$, similarity wiring is identical to no wiring and fully wiring, respectively. From the results shown in Figure \ref{fig:ex_ablation}(b), we observe that the best results can be obtained when $k=5$ and then the performance shows a decreasing trend with $k$ growth. These results suggest the significance of the selected initialization method.

\paragraph{Influence of Edge Diffusion Levels.}
We evaluate the sensitivity of the proposed method to different edge diffusion levels by comparing the LP performance with different teleportation probabilities of PPR $\alpha1, \alpha2 \in \{0.01, 0.05, 0.1, 0.2, 0.4\}$. The results are shown in Figure \ref{fig:ex_ablation}(c) and (d). We can observe that the proposed method performs better when the difference between $\alpha1$ and $\alpha2$ is larger. 
This result suggests that when contrasting the two views, a more distinguished diffusion difference between the two views makes richer topological information can be captured, improving the quality of learned node representations.

\paragraph{Influence of GNN Encoders.}
Our design enables 3SLP to be model-agnostic regarding the GNN encoder. Beyond the default GCN, we extend our evaluation of 3SLP's performance to three additional GNN encoders: SGC \cite{wu2019simplifying}, GIN \cite{xu2018powerful}, and GraphSAGE \cite{hamilton2017inductive}. 
As shown in Table \ref{table:gnn}, 3SLP exhibits consistently satisfying performance across different GNN encoders, demonstrating its compatibility. However, we notice that the performance with GraphSAGE is somewhat subpar, potentially due to the limitations inherent in its adopted neighborhood sampling strategy.

\paragraph{Influence of Similarity Metrics in PSC.}
Similarity metrics are pivotal in the PSC backbone, influencing the ultimate LP outcomes. In Table \ref{tab:NAO-LP}, we showcase the performance across all five introduced metrics for a comprehensive comparison.
The results show that the best metric varies from dataset, except the Manhattan distance, which consistently underperforms compared to others. Furthermore, our repetitive tests also suggest that the cosine distance metric achieves more consistently good performance, making it our preferable configuration.

\section{Discussion on Applicability} \label{sec:discussion}
In this work, the proposed method follows the notion of similarity-based NAO-LP and attempts to improve the baseline. The notion is based on the intuition that two nodes sharing similar features are more likely to be linked. However, not all graphs are highly consistent with this characteristic. Therefore, there exists a performance upper bound for similarity-based methods given the node attributes, which also indicates the limitation of 3SLP. 
Another factor is the scale of graph, which can influence the learning effect of the GNN encoder.
These facts are also reflected in the different performances of 3SLP on different datasets.
Nonetheless, we demonstrate the improvement brought by the SSL within such a scope of applicability.

\section{Conclusion} \label{sec:conclusion}
In this paper, we propose 3SLP to rejuvenate the similarity-based LP methods. 
We integrate self-supervised graph representation learning techniques into 3SLP to enable it to handle the realistic use case of similarity-based LP.
Without the supervision of edge labels, 3SLP can develop informative node representations as inputs instead of original node attributes to the pairwise similarity-based clustering backbone to prompt LP performance on attributed homophilic graphs. We empirically demonstrate the superiority of 3SLP compared to the baselines. 
Our extensive analysis highlights the crucial impact of graph homophily on the efficacy of 3SLP.
In future research endeavors, we aim to explore the applicability of the proposed method to heterophilic graphs.





\newpage

\bibliographystyle{ACM-Reference-Format}
\bibliography{ref}


\end{document}